\documentclass[letterpaper, 10 pt, conference]{ieeeconf}

\IEEEoverridecommandlockouts
\usepackage{booktabs}       
\usepackage{amsfonts}       
\usepackage{nicefrac}       
\usepackage{microtype}      
\usepackage{amsmath}
\usepackage{amssymb}

\usepackage{graphicx}
\usepackage{float}
\usepackage{booktabs}
\usepackage{ltxtable}
\usepackage{dblfloatfix}

\title{Context is Key: New Approaches to Neural Coherence Modeling}

\author{
David McClure$^{*}$, Shayne O'Brien$^{*}$, and Deb Roy%
\thanks{$^{*}$ indicates equal contribution.}%
\thanks{Shayne O'Brien is supported by an NSF Graduate Research Fellowship.}%
}

\begin{document}

\maketitle
\thispagestyle{empty}
\pagestyle{plain}
\pagenumbering{arabic}

\begin{abstract}

We formulate coherence modeling as a regression task and propose two novel methods to combine techniques from our setup with pairwise approaches. The first of our methods is a model that we call ``first-next,'' which operates similarly to selection sorting but conditions decision-making on information about already-sorted sentences. The second consists of a technique for adding context to regression-based models by concatenating sentence-level representations with an encoding of its corresponding out-of-order paragraph. This latter model achieves Kendall-tau distance and positional accuracy scores that match or exceed the current state-of-the-art on these metrics. Our results suggest that many of the gains that come from more complex, machine-translation inspired approaches can be achieved with simpler, more efficient models.

\end{abstract}

\section{Introduction}
The goal of coherence modeling, also known as sentence ordering, is to organize a given set of interdependent sequences into a coherent ordering. This task is typically modeled by randomly shuffling the sentences of a paragraph and then using some algorithm to try and correctly restore the document to its gold ordering. Despite its conceptual simplicity, this objective proves to be non-trivial to solve on corpora covering a diverse range of topics despite sensical text having higher-level structure.

Coherence modeling is an important task in many real-world applications such as  multi-document summarization, retrieval-based question answering, conversational analysis, topic modeling, automated text evaluation, and natural language generation. Two examples of the benefits of determining sentence order include enhancing user comprehension in human computer interaction applications and streamlining the grading of standardized test essays (Barzilay and Elhadad, 2002; Logeswaran et al., 2016). 

Given its potential real-world applications, coherence modeling has increasingly gained attention from natural language processing researchers since 2016. Significant work on this problem before this was primarily focused around centering theory (Grosz et al., 1995) and modeling based on: linguistic feature extraction (Lapata, 2003); global coherence using hidden Markov models to capture document structure (Barzilay and Lee, 2004); and local coherence gained from patterns of entity distributions (Barzilay and Lapata, 2008). The success of these approaches was limited to datasets that were relatively restricted in linguistic scope, namely the \texttt{ACCIDENTS} dataset of aviation accident reports and the \texttt{EARTHQUAKES} dataset of newspaper articles about earthquakes.

More recent work on coherence modeling has focused on ordering research paper abstracts, in particular the open-source arXiv abstracts dataset consisting of 1,106,141 abstracts from six different scientific disciplines. This dataset is introduced and summarized in Chen et al. (2016). Approaches to modeling these data have included a machine translation framework (Gong et al., 2016), computing pairwise prediction and using beam search to output the most likely ordering (Chen et al., 2016), and probabilistic text structuring via sentence graph dependency parsing (Li and Jurafsky, 2017).

With respect to the arXiv abstracts dataset, the contributions of this paper are as follows:
\begin{enumerate}
\item We formulate coherence modeling as a regression task by mapping sentences onto a continuous linear axis and sorting on predicted values. In the case where Kendall-tau distance is used as an evaluation metric, this setup avoids loss mismatch.

\item Building on the pairwise approach from Chen et. al (2016), we introduce the ``first-next'' model that combines local context gained from the pairwise models with global coherence provided by the regression models. This achieves near state-of-the-art performance on the perfect match ratio and Kendall-tau metrics.

\item We show that a simple modification to the regression paradigm to incorporate context matches or exceeds the best published results for the Kendall-tau and positional accuracy metrics. This model is simpler and more efficient than previously introduced models, suggesting that complex models are not needed to learn strong models for this task.
\end{enumerate}

\section{Evaluation Metrics}
We are given a corpus of $N$ sequences, each consisting of $n$ potentially out-of-order items $\{ s_1, \ldots, s_n \}$. The objective of coherence modeling is to approximate the gold ordering of a sentence $i$, denoted $o_{i}$, using 
\[ \hat{o}_{i} = \{ \mathbb{I} \{ \hat{s}_1 = 1 \}, \ldots, \mathbb{I} \{ \hat{s}_n = n \} \} \] 
where $\hat{s}_i$ is the predicted position of sentence $s_i$ and $\hat{o}^{j}_{i} =  \mathbb{I} \{ \hat{s}_j = j \}$ is the indicator function equal to 1 if $\hat{s}_j = j$ and 0 otherwise. Model evaluation is completed using three metrics commonly used in the literature: Kendall tau distance, perfect match ratio, and positional accuracy.

\subsection{Kendall Tau Distance}
The Kendall tau distance (KT) falls in the continuous interval [-1, 1] and measures the ordinal distance between two ordered sequences (Kendall, 1945). Formally,

\[ KT = \frac{P+Q}{P-Q} \]
where $P$ is the number of concordant pairs of positions between the gold and predicted ordering and $Q$ the number of discordant pairs. Given a prediction and its corresponding gold target the KT will be -1 if the predicted ordering is the reverse of the gold, near or equal to 0 if the predicted ordering is approximately random in comparison to the gold ordering, and 1 if the prediction ordering matches the gold ordering exactly. 
Lapata (2006) establishes that this evaluation metric reliably correlates with human judgments.

\subsection{Perfect Match Ratio}
The perfect match ratio (PMR) is defined as the ratio of exactly matching orders across all predicted sequences. Let $|\mathcal{S}|$ be the total number of documents in the test dataset. Then

\[ PMR = \frac{1}{N} \sum\limits_{i=1}^{N} \mathbb{I} \lbrace{\hat{o}_i = o_{i}}\rbrace \]
where $\mathbb{I} \lbrace{\cdot}\rbrace$ is the indicator function equal to 1 if  $\hat{o}_i = o_{i}$ exactly and 0 otherwise. 

\subsection{Positional Accuracy}
The positional accuracy (PA) is defined as the proportion of positions in the predicted ordering that match that of the gold ordering. In particular,
\[ PA = \frac{1}{|\mathcal{S}|} \sum\limits_{i=1}^{|\mathcal{S}|} \sum\limits_{j=1}^{|o_i|} \mathbb{I} \{ \hat{o}_{i}^j = o_{i}^{j} \} \]
where $|o_i|$ is the number of items in $o_i$, and $\mathbf{I} \lbrace{\cdot}\rbrace$ is the indicator function equal to 1 if $\hat{o}_{i}^j = o_{i}^j$ and 0 otherwise. Compared to PMR, PA is a less difficult metric to maximize as predictions that are only partially correct are still counted in the overall evaluation.

\section{Experimental Setup}

We formulate coherence modeling as a regression task in which gold tags for each sentence in the document are linearly mapped onto the continuous interval [-1, 1]. Consecutive tags have even numerical spacings that scale appropriately with the number of sentences in the sequence. Loss is computed with PyTorch's \texttt{MSELoss} function using predicted versus gold KT scores as input. We train for 150 epochs by randomly selecting 128 abstracts per batch for 100 steps per epoch. Early stopping based on validation KT score is used. We featurize each of the sentences in the abstracts at the word-level via Google News word embeddings with dimensionality $d$ = 300. Tokenization is done using the \texttt{word\_tokenize} function from the NLTK python package. Sentences are padded with $d$-dimensional vectors of zeros if their length after tokenization is less than 64, and truncated to length 64 if longer. A maximum length of 64 was chosen because this value corresponds to the 99.5th percentile for sentence length across all sentences in the dataset. In all baselines, we use the Adam optimizer with a weight decay of 1e-4 and learning rate of 1e-4. To arrive at a final prediction at inference-time, sequences of regression values are argument sorted according to value.

\section{Baselines}

\subsection{Standard Linear Regression}

To get a sense of how much the raw lexical content of the sentences is predictive of position in the paragraph, we train a bag-of-words linear regression model. To do this we first build a vocabulary of the 3,000 most frequent unigrams, the 2,000 most frequent bigrams, and the 1,000 most frequent trigrams in the training dataset.\footnote{Using the full vocabulary of the dataset was computationally prohibitive.} Using these counts, a sparse feature vector is produced for each input sentence that consists of the number of times that each word in the vocabulary appears in the sentence. We then fit a standard linear regression to map the sentences onto the real-valued interval $[-1, 1]$.

\subsection{Continuous Bag of Words}

We implement the continuous bag of words (CBoW) model described by Mikolov et al. (2013). In this model, the word embeddings of each of the $n$ words in a sentence $s_{i}$ are averaged across all dimensions to output go from $n$x$d$-dimensional array of concatenated word-level representations to a $1$x$d$-dimensional array sentence encoding. Formally,
\[ \textbf{e} = \frac{1}{n} \sum\limits_{i=1}^{n}\textbf{e}_i \]
where $\textbf{e}$ $\in$ $\mathbb{R}^{d_{s_{i}}}$ and $\textbf{e}_m$ $\in$ $\mathbb{R}^{d}$, and $d_s$ = $d$ are the dimensions of the sentence and word embeddings, respectively. After averaging, we pass the $d$-dimensional array into a multilayer perceptron consisting of two feedforward neural networks (FFNN) of hidden dimension size $h$ = 1024 and a \texttt{tanh} activation layer between them. The output of the second layer is mapped to a vector of scalar outputs that represent the regression value(s) of the prediction for the input sentence(s).

\subsection{Convolutional Neural Network}
We implement a convolutional neural network (CNN) as described by Simard et al. (2003). Sentences are represented as
\[ \textbf{conv}_k = \phi ({W_{conv}^T}(\odot_{u=0}^{{l_f}-1}\textbf{e}_{k+u}) + \textbf{b}_{conv}) \]
\[ \textbf{e} = max_k \textbf{conv}_k\]
where $\textbf{W}_{conv}$ $\in$ $\mathbb{R}^{(d\times l_{f})\times d_{f}}$ and $\textbf{b}_{conv}$ $\in$ $\mathbb{R}^{d_{f}}$ are trainable parameters, and $\phi(\cdot)$ is the \texttt{tanh} activation function. In our case, $k$ = 1, $\ldots$, $|m|$ - $l_f$ + 1, and $l_f$ and $d_f$ are hyperparameters for the filter length and number of feature embedding maps, respectively.

We let $l_f$ = 3 and $d_f$ = 1024. Additionally, we add a $1 \times d$-dimensional vector of zeros on both sides of the input length. After padding, we pass the input first into the convolution layer and then into a \texttt{tanh} activation function. Maxpooling is completed on the output of this activation function over $k$, thereby reducing the last input dimension to 1. The output of this layer is then used as input for a FFNN with $h$=1024 that maps it to a scalar that represents the regression value of the input. This implementation is similar to that of Chen et al. (2016), with the only difference being our chosen value for $d_f$.

\subsection{Long Short-Term Memory Neural Network}

Formally, the long short-term memory (LSTM) neural network of Hochreiter and Schmidhuber (1997) is a more intricate version of the traditional recurrent neural network. LSTMs have memory cells $\textbf{c}$ $\in$ $\mathbb{R}^{d_f}$ modulated by three kinds of gates: input gates $\textbf{i}$ $\in$ $\mathbb{R}^{d_f}$, forget gates $f$ $\in$ $\mathbb{R}^{d_f}$, and output gates $\textbf{k}$ $\in$ $\mathbb{R}^{d_f}$. With $h_t$ denoting the hidden state at time step $t$, the LSTM update equation is given by:
\[ h_t, c_t = \texttt{LSTM}(h_{t-1}, c_{t-1}) \]
where
\[i_t = \sigma (W_i h_{t-1} + b_i) \] 
\[f_t = \sigma (W_f h_{t-1} + b_f)\]
\[k_t = \sigma (W_k h_{t-1} + b_o)\]
\[\hat{c}_t = \texttt{tanh} (W_c h_{t-1} + b_c)\]
\[c_t = f_t \odot (c_{t-1} + i_t \odot \hat{c}_t)\]
\[h_t = k_t \odot \texttt{tanh} (c_t)\]
and $W_{\lbrace h_i, h_f, h_k, h_c \rbrace }$, $W_{ \lbrace x_i, x_f, x_k, x_c \rbrace }$, and $b_{ \lbrace i, f, k, c \rbrace}$ are learnable parameters.

We let the the number of cells in the LSTM be $h$=1024. Similarly to the CNN, this is the only difference in the architecture compared to the implementation from Chen et al. (2016).

\section{Pairwise Ranking Model}

These regression architectures are limited by the fact that they only look at individual sentences in isolation. This means that tend to learn simple but dominant features that consistently show up in sentences that appear in particular positions in paragraphs, e.g. that "first" often appears in the first sentence and "finally" in the last. As a first step toward incorporating paragraph-level context, we replicated the technique described by Chen et al. (2016), which formulates the problem as a pairwise ranking task with zero-one loss. An example with 0 loss is given by
\[\{s_1, s_2\} \rightarrow 1\]
and an example with 1 loss is formed by reversing the order, i.e.
\[\{s_2, s_1\} \rightarrow 0\]
As an aside, we note that an \textit{exactly} reversed sentence pair is only one of many possible ordering mistakes that the model could make. For example, it would be wrong to have $\{s_4\}\rightarrow \{s_3\}$, but also $\{s_4\}\rightarrow\{s_6\}$, $\{s_4\}\rightarrow \{s_1\}$, etc. To explore the limitations of this setup, we experimented with a second version of it in which the negative cases are formed by randomly choosing a second sentence from among all other sentences in the paragraph except for the correct one. Instead of training the model to identify whether two adjacent sentences are in the correct or incorrect order, we thus train it to differentiate between correctly and incorrectly sequenced pairs. We found that models trained on this setup learned less quickly and performed worse than the simpler pairwise training regime.

Once the sentence encoder and classifier are trained, they can be used to assign a score to any given ordering of sentences by summing up the individual transition scores for each pair of sentences in the ordering. For short paragraphs of up to five sentences in length, it is possible to exhaustively check all possible sequences. Since the number of permutations grows with $O(n!)$ where $n$ is sequence length, greedy decoding becomes computationally infeasible for longer paragraphs. We followed Chen et al. (2016) in using an $L$-length beam search to look for a sequence with a high aggregate score across all of the sentence transitions. Although Chen et al. used a 128-length beam search, we found no difference in results by using a 100-length beam search.

On the test set, our implementation produced a perfect PMR of 32.23\% for the test dataset, just shy of the 33.34\% reported by Chen et al. It is interesting to note that the Kendall-tau score for this model is just 0.54, which was slightly worse than the score of 0.55 achieved by our simple bag-of-words linear regression model. For abstracts of 2-3 sentences, the pairwise model does better than the linear regression baseline, with an average KT score of 0.85, compared to 0.78 for the linear regression. As the paragraph length increases, the pairwise model falls off more quickly with an average KT of 0.30 at sequences of length 10 compared to 0.40 for the regression setup. The sequential model does a better job at giving very high accuracy for short paragraphs where local accuracy is most important, but the regression model does better at giving a broad-strokes ordering for longer paragraphs.

\section{First-Next: Combining local context with global structure}

In an effort to combine local sensitivity to sentence transitions with high-level trends in paragraph structure, we propose a model that we call the ``first-next.'' Starting with an unordered set of sentences, this model works like a selection sort by picking the first sentence from among the remaining sentences (the ``right'' context). When making this selection at any given time step, though, the model also incorporates information about the sentences that would precede it to its ``left.'' In this way, the model tries to pick the sentence that should come \textit{first} given the content of the paragraph in which it is found, and then sentences that should come \textit{next} given information provided by already selected sentences. In this way, the model is designed to be able to learn paragraph-level trends in addition and also higher-order structure by learning to choose sentences that coherently follow the already-selected sentences.

At any given time-step, the ``first-next'' model assembles a feature vector consisting of these components for each candidate sentence in the unordered set:

\begin{enumerate}
\item A vector representing the candidate sentence.
\item A vector representing the last sentence in the ``left'' context ($n$-1), the sentence that was chosen by the last time step.
\item A vector representing the $n$-2 sentence. Together, these effectively represent a ``trigram'' sentence model.
\item The offset position inside the paragraph of the current time-step. For example, if three sentences have already been chosen (indexes 0-2), the index of the current time-step is 3.
\item The offset position as a ratio, normalized between 0 and 1.
\item The randomly shuffled sentences in the "right" context, encoded as a single vector with an LSTM.
\end{enumerate}

During training, multiple positive and negative examples are generated for each abstract: for each $\{i\}^{n-1}_{i=1}$, a positive example is formed from the correct next sentence from the right context, and a negative example is formed by randomly picking a sentence from the right context that is not the next sentence. In total, three components are trained: an LSTM that encodes words in sentences, an LSTM that encodes shuffled sentences from the ``right'' context at any given time-step, and a classifier that makes a binary decision given the input vector described above. To generate an order for a shuffled abstract, we use a simple greedy decoding procedure, allowing the model to take the highest-scoring sentence at each time step. As future work, this could likely be improved by implementing a beam search similar to the one used in the pairwise model.

Using the greedy decoder, the first-next model produced a PMR performance of 36.05\% and Kendall-tau score of 0.69 on the test dataset. These represent a 4\% and a 15\% improvement over the pairwise ranking model on these metrics. Comparing the KT scores for different paragraph lengths for the two models, it is clear that our ``first-next'' model retains the high accuracy of the pairwise ranking model for short abstracts as well as the relatively high accuracy of the regression model for the longer abstracts.

\begin{table*}[!htb]
  \centering
  \begin{tabular}{c c c c} \toprule
    Model & Perfect Match Ratio & Kendall tau & Positional Accuracy \\ \toprule
    CBOW regression & 0.179 & 0.399 & 0.314 \\ \midrule
    Linear regression & 0.239 & 0.549 & 0.393 \\ \midrule
    LSTM regression & 0.241 & 0.546 & 0.396 \\ \midrule
    CNN regression & 0.268 & 0.603 & 0.427 \\ \midrule
    Pairwise Ranking Model & 0.322 & 0.544 & 0.407 \\ \midrule
    First-Next & 0.360 & 0.688 & 0.514 \\ \midrule
    Regression with context & \textbf{0.362} & \textbf{0.714} & \textbf{0.527} \\ \bottomrule
  \end{tabular}
  \caption{Final results for all models} \label{tab:results}
\end{table*}

\section{Adding context to the regression paradigm}
Since the ``first-next'' model seems to add information about the global paragraph context to the pairwise approach, we were curious if it would be helpful to add global context to the original regression baselines. Instead of just encoding an individual sentence and regressing it onto a continuous linear axis, we now encode the entire, shuffled paragraph context as a single vector and concatenate this with the sentence vector before performing the regression. We also include a single extra dimension that explicitly provides the number of sentences in the paragraph, which seems to slightly help the model make decisions about where to place sentences. These combined vectors are then regressed onto a linear axis using a FFNN. Like before, these regression predictions can then be used as sorting keys for the sentences.

Our context-aware model achieved a PMR of 36.2\% on the test abstracts (narrowly beating the ``first-next'' model) and an average KT score of 0.714, representing a 2.4\% improvement over first-next. Its KT performance comes close to state-of-the-art performance, though it is difficult to compare directly to other results since different testing corpora are used. In particular, Logeswaran et al. (2016) achieve KT scores of 0.72 and 0.73 on corpora of NIPS and AAN abstracts using a many-to-one sequence model inspired by Vinyals et al. (2015), but these corpora are very small: just 402 and 2,626 test abstracts, respectively. When tested on a larger corpus of 127,835 NSF abstracts, their model achieved a Kendall-tau score of only 0.51. Meanwhile, the perfect accuracy score falls short of the results reported by Gong et al. (2016) using the same corpus and a model similar to the one used by Logeswaran et al. (2016). Unfortunately, Gong et al. (2016) did not report KT scores.

In addition to our simple contextual regression model achieving the strongest performance on all three metrics, it is also worth noting that this model has a number of desirable characteristics from a more pragmatic, engineering standpoint. It trains faster than any other models we tested and it can produce order predictions very quickly since it does not need any kind of decoding algorithm. The regression outputs can be used as sorting keys for the sentences, meaning that all of the computationally expensive work is done upfront and streamlined by the sentence and context encoders.

\section{Discussion and future work}

We trained five baseline classes of models: linear regression, continuous bag-of-words, CNN, LSTM, and pairwise ranking. We then proposed the ``first-next'' model to better capture local and global indicators of coherence. Finally, we combine the strengths of the pairwise and regression approaches using our context-aware regression model. After experimenting extensively with different encoding architectures, hyperparameters, and training designs for all models, we found this latter model to perform best. Despite the relative simplicity of regression approaches, they appear to be powerful for ordering tasks involving text. Our findings seem to suggest that the performance gains that have come from using more complex set-to-sequence and machine-translation-inspired techniques are attainable by simpler, more efficient models.

This being said, we do not claim to have a nuanced understanding at this point of why the regression-with-context model works. It seems to us that the addition of the context vector allows the model to pick up on relationships between the individual sentence and the content of the rest of the paragraph, which allows it to make more accurate regression predictions and do a better job of sorting the sentences in the entire paragraph. Still, it is not obvious what types of interactions between the local and global context are most salient. 

In future work, we hope to investigate this further, and come up with a more interpretable account of what the model is learning. Additionally, we are interested in using this result as the jumping-off point for another line of work that we spent quite a bit of time thinking about while working on this project. Instead of approaching coherence modeling as an ordering task in the traditional sense, we are interested in the idea of training models that can act as ``fitness functions.'' This is meant to mean models that would look at an entire paragraph and produce an estimation of the degree to which it is correctly ordered. Unlike previous work on coherence modeling, though, the goal would be to estimate a continuous score--perhaps a Kendall-tau value--instead of just treating the problem as a binary classification task that decides between correct and incorrect orderings.

With this in hand, we believe it would be possible to use a genetic approach to search for an optimal ordering similarly to the randomized greedy approach applied to dependency parsing in Zhang et al. (2014). Instead of starting with a totally random ordering, we believe it could be useful to start with a reasonably good estimate of the order and make selective, surgical changes. We wonder if the contextual regression model could be a good starting point for this given that it is simple to train, fast to generate predictions, approaches or exceeds start-of-the-art performance, and does a good job of producing reasonably good orderings even for longer paragraphs.

\begin{figure*}
  \centering
  \includegraphics[width=0.7\textwidth]{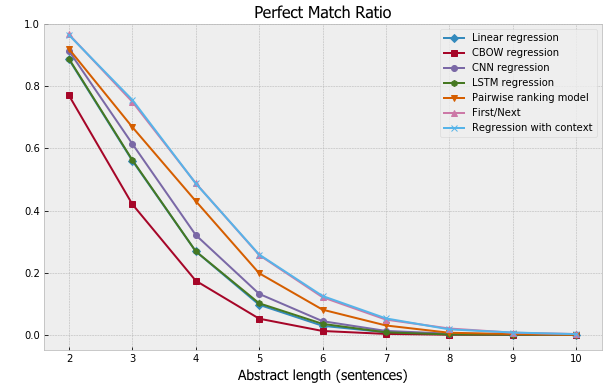}
  \caption{On the PMR metric, almost all of the gains for the higher-performing models come from abstracts with fewer than 7 lines, after which the task becomes extremely difficult.}
\end{figure*}

\begin{figure*}
  \centering
  \includegraphics[width=0.7\textwidth]{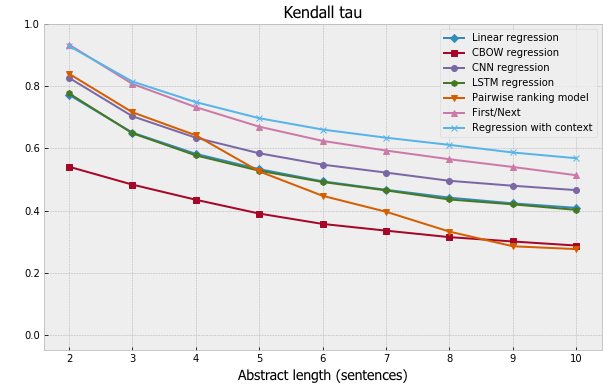}
  \caption{As the paragraph length increases, the regression and first-next models do a good job of preserving a roughly correct ordering. Whereas, the pairwise ranking model falls off quickly after 4-5 sentences. It also seems as if the performance increase for the contextual regression model over the first-next model comes mainly from better performance on the longer abstracts.}
\end{figure*}



\begin{figure*}
  \centering
  \includegraphics[width=0.7\textwidth]{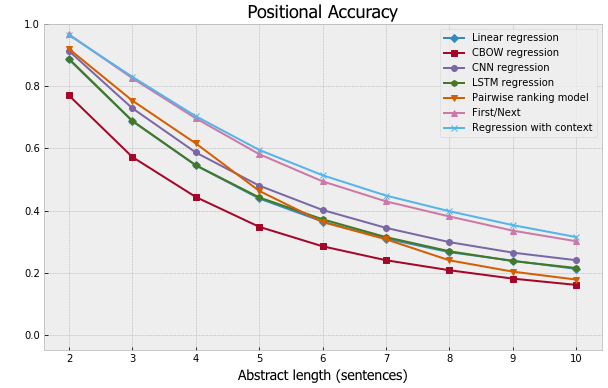}
  \caption{As the paragraph length increases, the regression and first-next models do a good job of preserving a roughly correct ordering. Whereas, the pairwise ranking model falls off quickly after 4-5 sentences.}
\end{figure*}


\begin{thebibliography}{99}

\bibitem{c1} R. Barzilay and N. Elhadad. Inferring Strategies for Sentence Ordering in Multidocument News Summarization. Journal of Artificial Intelligence Research 17:35-55, 2002.

\bibitem{c2} R. Barzilay and L. Lee. Catching the Drift: Probabilistic Content Models, with Applications to Generation and Summarization. HLT-NAACL. 2004.

\bibitem{c3} R. Barzilay and M. Lapata. Modeling Local Coherence: An Entity-Based Approach. Computational Linguistics, 34(1):1–34, 2008.

\bibitem{c4} X. Chen, X. Qiu, and X. Huang. Neural Sentence Ordering. arXiv preprint arXiv:1607.06952. 2016.

\bibitem{c5} J. Gong, X. Chen, X. Qiu, and X. Huang. End-to-End Neural Sentence Ordering Using Pointer Network. arXiv preprint arXiv:1607.06952v2. 2016.

\bibitem{c6} B. J. Grosz, S. Weinstein, and A. K. Joshi. Centering: A Framework for Modeling the Local Coherence of Discourse. Computational Linguistics, 21(2):203–225, 1995.

\bibitem{c7} S. Hochreiter and J. Schmidhuber. Long Short-Term Memory. Neural Computation 9(8): 1735-1780. 1997.

\bibitem{c8} M. Kendall. The Treatment of Ties in Ranking Problems. Biometrika 33(3): 239-251. 1945.

\bibitem{c9} M. Lapata. Automatic evaluation of information ordering: Kendall’s tau. Computational Linguistics, 32(4):471–484, 2006.

\bibitem{c10} M. Lapata. Probabilistic Text Structuring: Experiments with Sentence Ordering. In Proceedings of the 41st Annual Meeting on Association for Computational Linguistics-Volume 1, pages 545–552. Association for Computational Linguistics, 2003.

\bibitem{c11} J. Li and D. Jurafsky. Neural Net Models of Open-domain Discourse Coherence. In proceedings Empirical Methods in Natural Language Processing, pp. 198-209, 2017.

\bibitem{c12} L. Logeswaran, H. Lee and D. Radev. Sentence Ordering Using Recurrent Neural Networks. arXiv preprint arXiv:1611.02654. 2016.

\bibitem{c13} T. Mikolov, K. Chen, G. Corrado, and J. Dean. Efficient Estimation of Word Representations in Vector Space. ICLR Workshop Papers. 2013.

\bibitem{c14} P. Simard, D. Steinkraus, and J. Platt. Best Practices for Convolutional Neural Networks Applied to Visual Document Analysis. ICDAR. 2003.

\bibitem{c15} O. Vinyals, M. Fortunato, and N. Jaitly. Pointer Networks. arXiv preprint arXiv:1611.02654v2 [stat.ML]. 2015.

\bibitem{c16} Y. Zhang, R. Barzilay, and T. Jaakkola. Greed is Good if Randomized: New Inference for Dependency Parsing. EMNLP. 2014.

\end{thebibliography}
\end{document}